\documentclass[journal]{IEEEtran}
\IEEEoverridecommandlockouts
\usepackage{cite}
\usepackage{amsmath,amssymb,amsfonts}
\usepackage{algorithmic}
\usepackage{graphicx}
\usepackage{textcomp}
\usepackage{xcolor}
\usepackage{bm}
\usepackage{lineno,hyperref,float,amsfonts,amsmath,epsfig,graphics,booktabs,multirow,cite,stfloats}
\usepackage{amsthm,mathrsfs}
\usepackage{url}
\usepackage{colortbl}
\usepackage{subfigure}
\usepackage{comment}

\usepackage{threeparttable}

\theoremstyle{definition}

\theoremstyle{plain}

\theoremstyle{remark}

\newcommand{\tabincell}[2]{\begin{tabular}{@{}#1@{}}#2\end{tabular}}

\def\BibTeX{{\rm B\kern-.05em{\sc i\kern-.025em b}\kern-.08em
    T\kern-.1667em\lower.7ex\hbox{E}\kern-.125emX}}
\begin{document}

\title{AI and Open-data Driven \\Scalable Solar Power Profiling}
\author{Shiliang Zhang, Sabita Maharjan, Damla Turgut
\thanks{This work was supported by UiO:Energy Convergence Environments via the PriTEM project. Shiliang Zhang and Sabita Maharjan are with the Department of Informatics, University of Oslo, Norway (e-mail: \{shilianz, sabita\}@ifi.uio.no). Damla Turgut is with the Department of Computer Science, University of Central Florida, USA (e-mail: damla.turgut@ucf.edu).}
}

\maketitle

\begin{abstract}
Solar photovoltaic (PV) deployment is expanding rapidly, yet detailed, up-to-date information on the spatial distribution and capacity of rooftop PV remains limited. This paper presents an open, scalable framework for detecting solar panels from open data and generating city-level solar power profiles. We leverage foundation vision AI models to detect solar panel geometries from open-source satellite imagery. This avoids manual data labeling and case-specific model training while maintaining robustness across heterogeneous imagery. Detected solar panels are converted into georeferenced polygons, yielding spatially explicit and incrementally extensible inventories. By integrating open weather data, we translate panel footprints into regional solar power profiles. The framework reduces dependency on proprietary imagery, manual labeling, and closed‑source models, and offers a transparent and scalable approach for solar planning and analysis. We released the data and an API resulted from this work. For any user‑specified building location, our API retrieves aerial imagery, detects rooftop solar panels, and returns georeferenced polygons. This empowers researchers and developers to scan user‑defined areas to build solar panel maps and associated solar production profiles, thus facilitating advanced analysis like distributed solar production integration, local power flow optimization, energy tariff design, and infrastructure planning.
\end{abstract}

\begin{IEEEkeywords}
AI, open data, solar panel detection, solar power profiling.
\end{IEEEkeywords}

\section{Introduction} \label{section1}

The growing deployment of solar photovoltaic (PV) systems is reshaping electricity systems worldwide. Accurate and timely information on the location and production capacity of distributed solar installations is critical for distribution grid operation, capacity planning, and urban energy modeling~\cite{li2025global,li2025transformer}. However, comprehensive solar panel inventories are rarely available and, where they exist, are often proprietary, incomplete, or difficult to reproduce~\cite{awadallah2025road,nyangiwe2026performance}. This has motivated automatic methods for solar panel detection from remote sensing data~\cite{barraz2025fast} and for deriving spatial and temporal solar power profiles~\cite{rao2025development}.

A common strategy is to train task-specific computer vision models on labeled satellite images~\cite{boccalatte2025leveraging}. While accurate, this requires substantial data collection and manual labeling~\cite{sezer2021detection} and incurs significant computational cost~\cite{cardoso2024automated}. Such models often generalize poorly to new regions, resolutions, or sensor characteristics. Reusing existing trained models~\cite{park2023boost} avoids retraining but ties performance to original data regimes. Differences in aerial image resolution, tiling schemes, acquisition angles, and spectral characteristics can degrade accuracy of pretrained models~\cite{yang2025large}. Adapting these models typically demands additional fine‑tuning data and expertise, limiting their usability as off‑the‑shelf detectors.

Recent advances in AI have prompted interest in large language models (LLMs) and multimodal variants for solar panel detection~\cite{guo2025solar}. However, LLMs are optimized for tokenized representations rather than dense, pixel‑level tasks~\cite{pmlr-v267-shabbir25a}. Textual outputs hinder spatial precision and downstream geographic processing and integration with energy system models~\cite{hu2026extracting}. For high‑fidelity solar panel detection, vision‑centric models remain more suitable~\cite{amorim2025collaborative}. Yet purely model‑centric pipelines can be unscalable~\cite{10.1145/3711118}. That is, the outputs are often pixels without explicit geographic location or extent~\cite{bradbury2016distributed}, limiting transferability and rendering possibly repeated computations by different users for the same area.

\begin{figure}[htbp]
    \centering
    \includegraphics[width=0.98\linewidth]{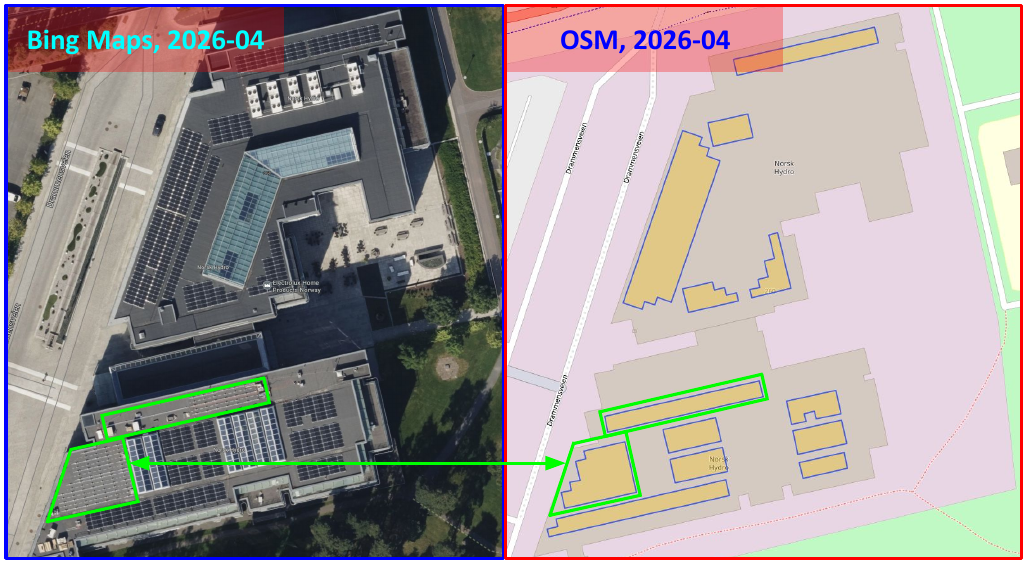}
    \caption{Example of solar panel polygons (right) from OSM and the corresponding aerial image (left). Panels labeled in OSM but not visible in Bing Maps (lower-left) indicate outdated tiles and discrepancies between aerial imagery and OSM data.}
    \label{fig:OSM_solar_panel_label_example}
\end{figure}

Open geographic data sources such as OpenStreetMap (OSM) provide georeferenced polygons that indicate solar panel locations and extents~\cite{11364211}. Where available, such labels are valuable (see Figure~\ref{fig:OSM_solar_panel_label_example}), but the coverage and quality vary across regions and follow mapping activity rather than actual deployment. OSM alone therefore cannot provide comprehensive solar inventories. Meanwhile, open imagery sources like JOSM~\cite{deri2025crowdsourced} and OpenAerialMap~\cite{mandourah2025analysing} offer opportunities to calibrate and complement OSM solar data.

In this work, we combine open data with generalizable AI models to detect solar panels and derive solar power profiles. Foundation AI models, trained on large and diverse datasets, can transfer across tasks and domains. We exploit this to perform solar panel detection on open aerial imagery, avoiding project‑specific data processing, labeling, and training and enhancing reproducibility.

\begin{figure*}[!b]
    \centering
    \includegraphics[width=0.9\linewidth]{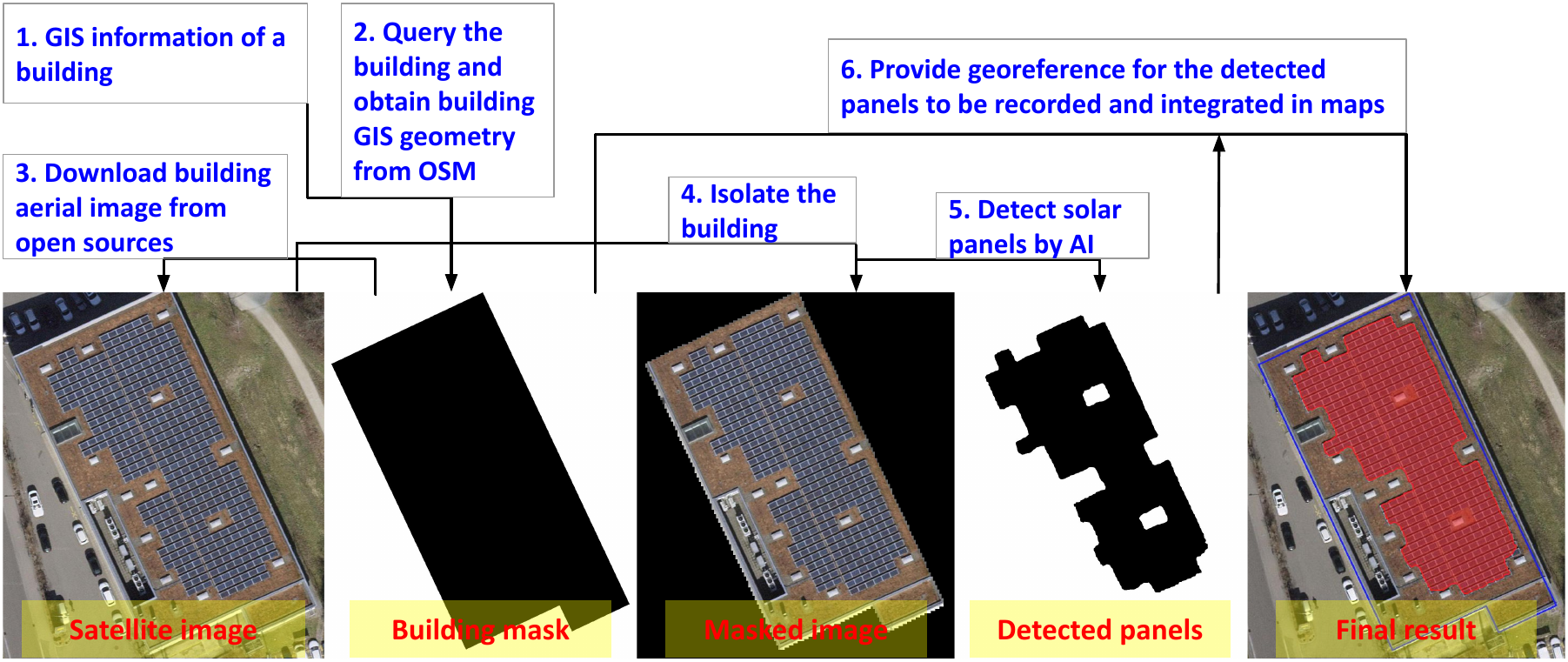}
    \caption{Workflow for solar panel detection and geolocation for a single building. GIS denotes geographic information system.}
    \label{fig:workflow}
\end{figure*}

To make detected solar panels spatially explicit and reusable, we align AI model outputs with open geographic information system (GIS) data from OSM. We represent detected panels as georeferenced polygons, enabling accumulation into an open solar panel inventory that supports rooftop potential estimation, building‑level attribution, and cross‑referencing with other datasets. We further incorporate open weather data to convert detected panels into time‑varying solar power profiles for regional solar resource assessment and planning. Furthermore, we release an API that, given a user‑specified building location, it retrieves imagery, applies the AI model, and returns georeferenced rooftop solar panel polygons. This empowers energy stakeholders to scan areas of interest, construct solar panel maps, and derive solar power profiles. We summarize the main contributions below:

\begin{itemize}
    \item[(i)] We develop an AI‑augmented framework for solar panel detection from open data leveraging open aerial imagery, thus releasing the need for case‑specific model training and proprietary data.
    \item[(ii)] We design a method to link detected solar panels with OSM. This leads to georeferenced solar panel representations that are reproducible, scrutinisable, and incrementally extensible as open data.
    \item[(iii)] We propose an approach to transform detected panel footprints into regional solar power profiles using open weather data, integrating visual detection with energy system‑relevant metrics.
    \item[(iv)] We publicly release an API for building‑level and area‑wide solar panel detection, combining open imagery, open GIS, and generalisable AI models for scalable solar panel mapping and power profiling.
\end{itemize}

We organize the paper as follows. Section~\ref{sec:detection} presents the AI-augmented solar panel detection from open geographic data. In Section~\ref{sec:profiling}, we convert detected panel pixels into georeferenced footprints and solar power profiles. Section~\ref{sec:evaluation} evaluates the approach in two areas in Switzerland. Section~\ref{conclusions} concludes the paper.

\section{AI and open-data driven solar panel detection}\label{sec:detection}

This section presents the detection pipeline based on open aerial imagery. We use OSM to anchor the analysis and retrieve imagery, conduct solar panel detection using foundation AI models, and associate the detection results with georeference. Figure~\ref{fig:workflow} outlines the detection workflow in this section. 

\subsection{Geospatial anchoring and imagery retrieval}

\begin{figure}[htbp]
    \centering
    \includegraphics[width=0.9\linewidth]{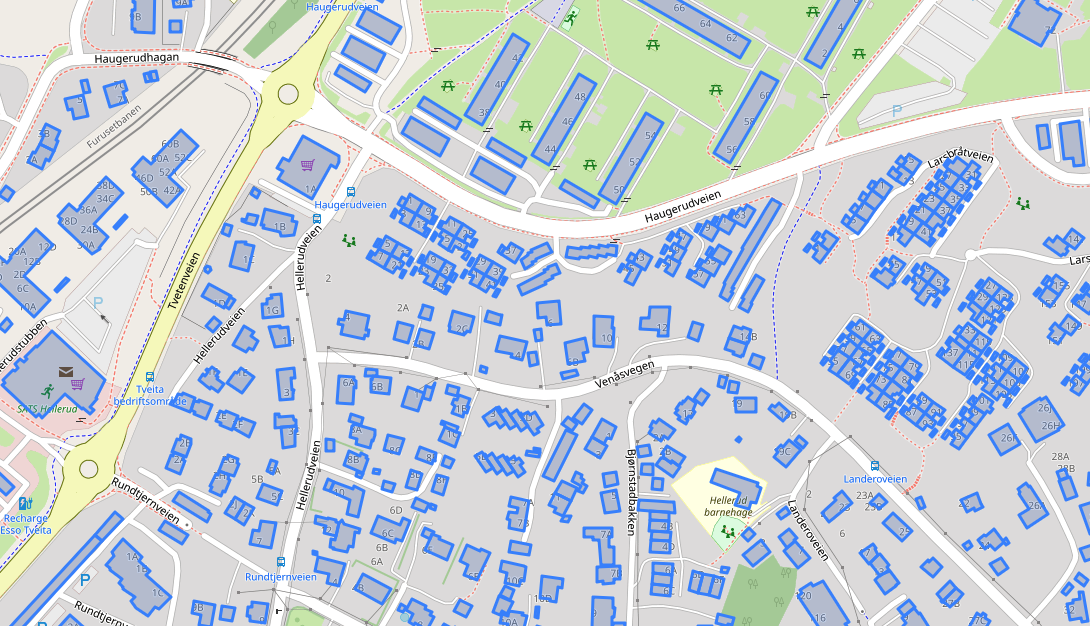}
    \caption{Example of building geometries retrieved from OSM.}
    \label{fig:Buildings}
\end{figure}

We use OSM as the primary reference for building rooftops. We obtain building geometries by querying OSM for the area of interest. OSM provides georeferenced polygons representing building footprints as shown in Figure~\ref{fig:Buildings}. These footprints define candidate rooftops and support further geolocation of detected solar panels.

\subsection{Open aerial imagery}

\begin{figure}[htbp]
    \centering
    \includegraphics[width=0.9\linewidth]{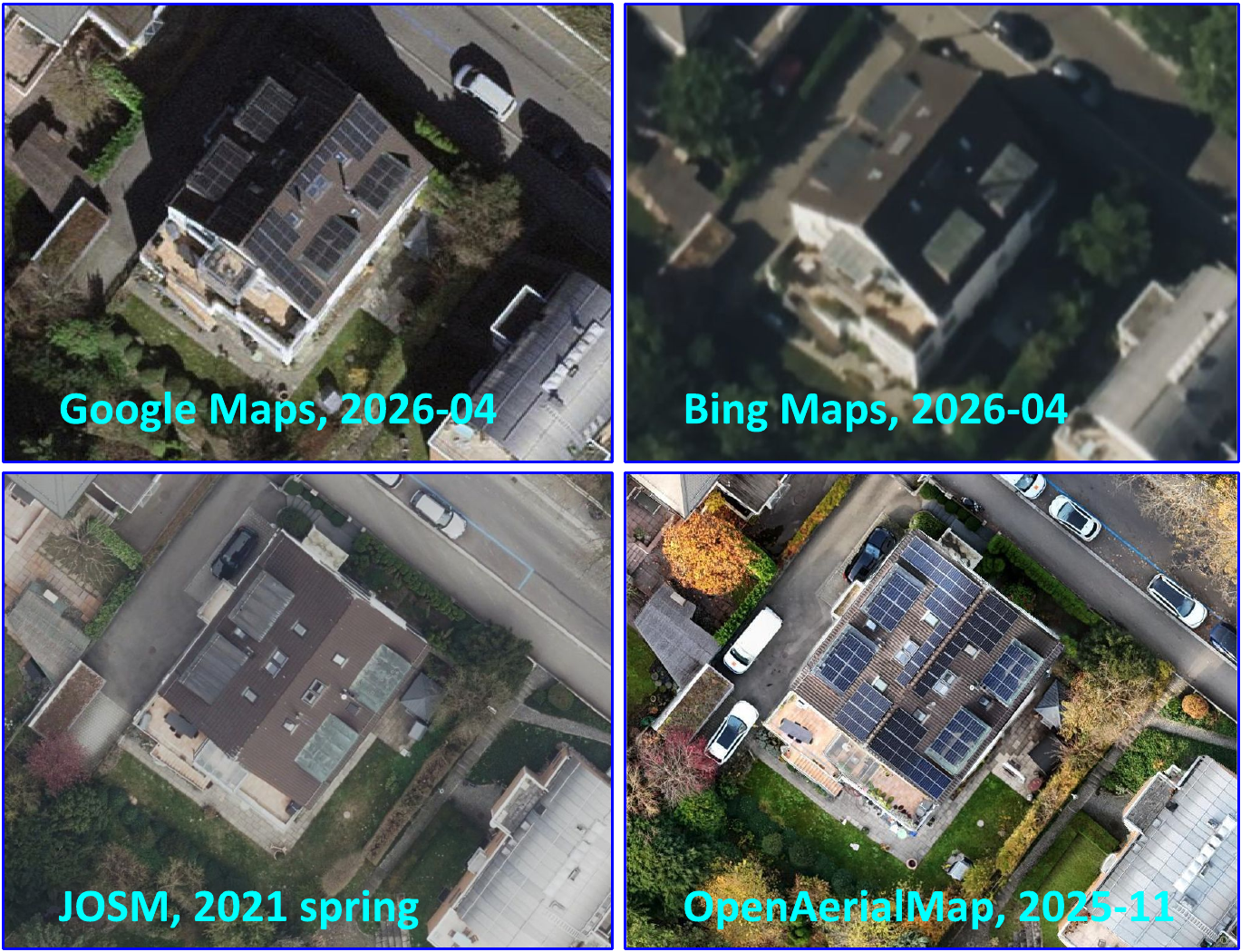}
    \caption{Sources of aerial images and example imagery for the same building in Zürich. Google and Bing Maps are proprietary. JOSM and OpenAerialMap provide open imagery.}
    \label{fig:Image_data_sources}
\end{figure}

Given building geometry, we retrieve aerial imagery from open sources. Historically, most aerial images were proprietary and licensed (\textit{e.g.}, Google Maps, Bing Maps). Open communities and UAV-based imagery now provide high‑resolution, timely images. Figure~\ref{fig:Image_data_sources} compares proprietary and open imagery, illustrating that in some areas, open sources imagery outperforms proprietary ones. We would like to note that open imagery sources such as JOSM\footnote{\url{https://josm.openstreetmap.de/wiki/Maps}, accessed 2026-04-26.} cover large regions and support city-level solar panel detection. Moreover, JOSM provides API for direct retrieval of images for given coordinates, enabling linkage between imagery and GIS information. In this paper, we use OSM building geometries and request aerial images covering each building from open imagery. This provides rooftop images and associated geographic information (latitude, longitude), which we use for georeferencing detected solar panels in Section~\ref{sec:profiling}.

\subsection{A foundation AI model based approach}

With building aerial images, we adopt foundation vision AI models for solar panel detection, leveraging their robustness to heterogeneous imagery. Foundation AI models (\textit{e.g.}, \textit{sam3} by Facebook\footnote{\url{https://github.com/facebookresearch/sam3}, accessed 2026-04-26.}) are trained on large and diverse datasets and exhibit strong generalization. We use such models to segment solar panels at pixel level from aerial images.

Our approach anchors inputs with OSM and open imagery, yielding an open-data-centric solution applicable across cities. We post-process detected solar panels, initially as pixels, and associate them with georeferences. This makes the detection result recordable, transferable, and directly integrable into maps, which enables scalable solar panel scanning and solar power profiling using only open data.

\section{Postprocessing of the detection results and solar power profiling generation}\label{sec:profiling}

\subsection{Georeference of the detected solar panel pixels}

\begin{figure}[!b]
    \centering
    \includegraphics[width=0.9\linewidth]{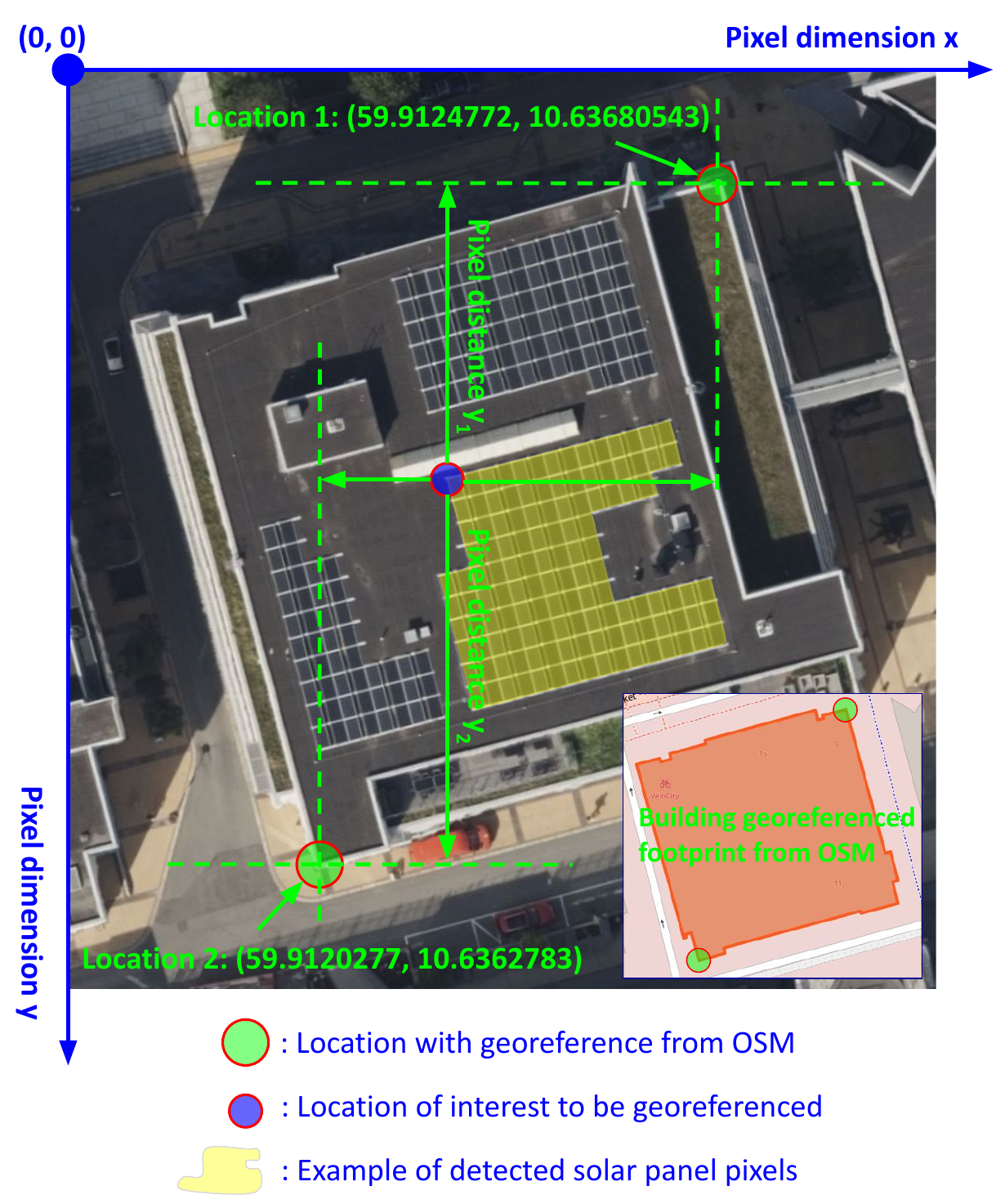}
    \caption{Example of georeferencing solar panel pixels for a single building.}
    \label{fig:georeference}
\end{figure}

We derive a linear interpolation approach to determine geographic coordinates of solar panels detected within aerial imagery. As illustrated in Figure~\ref{fig:georeference}, we use reference locations from OSM building footprints to georeference detected panels.

In Figure~\ref{fig:georeference}, we use two reference locations $L_1$ and $L_2$. The image origin is at $(0,0)$ in the upper-left. The geographic coordinates are $G_1 = (\text{lat}_1, \text{lon}_1)$ and $G_2 = (\text{lat}_2, \text{lon}_2)$. The point of interest $P_t$ is a pixel within the detected solar panel areas, located between $L_1$ and $L_2$ with vertical pixel distances $y_1$ from $L_1$ and $y_2$ from $L_2$. The total vertical pixel displacement is
\begin{equation}
    \Delta Y_{pixel} = y_1 + y_2.
\end{equation}
Assuming a linear transformation between pixel grid and geographic plane, the latitude of $P_t$ is
\begin{equation}
    \text{lat}_t = \text{lat}_1 + \left( \frac{y_1}{y_1 + y_2} \right) \cdot (\text{lat}_2 - \text{lat}_1).
\end{equation}
Similarly, the longitude is
\begin{equation}
    \text{lon}_t = \text{lon}_1 + \frac{\Delta x_1}{\Delta X_{pixel}} \cdot (\text{lon}_2 - \text{lon}_1),
\end{equation}
where $\Delta x_1$ is the horizontal pixel distance from $P_t$ to $G_1$, and $\Delta X_{pixel}$ is the total horizontal pixel distance between $G_1$ and $G_2$. Applying this to all pixels covered by detected solar panels yields their geolocations, from which we extract panel polygons and areas and build extensible solar panel inventories.

\subsection{Solar power profiling based on open weather data}

We generate hourly annual time series of solar power output for solar panels at a given location using meteorological and PV performance models. We first retrieve the GIS information of the city. Given a city name, we obtain its coordinates via an online geocoding service. Let $\phi$ and $\lambda$ denote latitude and longitude.
\begin{figure}[htbp]
    \centering
    \includegraphics[width=0.9\linewidth]{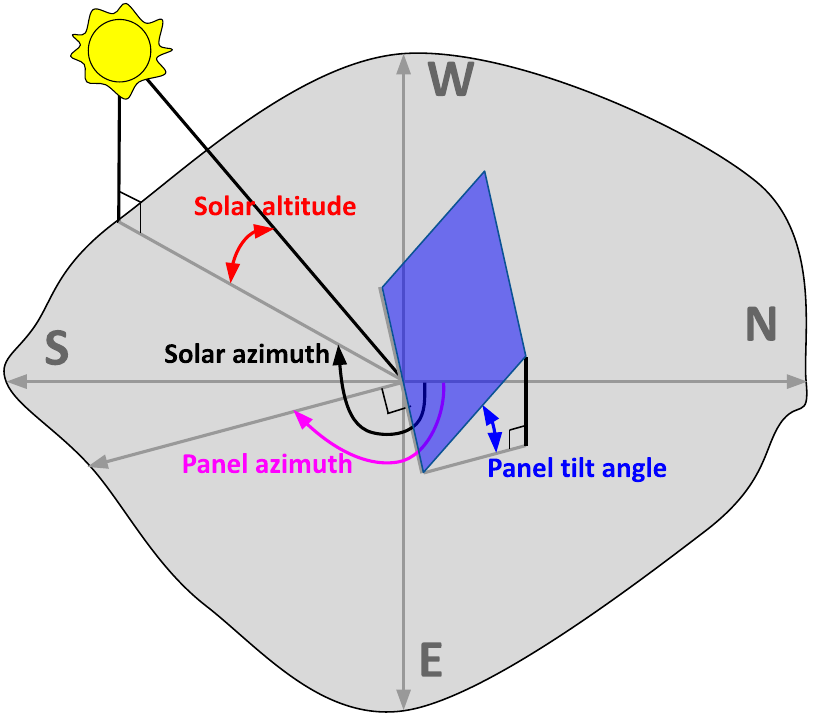}
    \caption{Relationship between sun irradiation and solar panel orientation.}
    \label{fig:sun_panel}
\end{figure}

We compute the apparent solar position at hourly steps over one year:
\begin{equation}
  \{t_i\}_{i=1}^{N}, \qquad
  N = 365 \times 24 = 8760,
\end{equation}
with one timestamp per hour. For each $t_i$, we calculate solar altitude $\alpha_i$ and azimuth $\gamma_{s,i}$, as illustrated in Figure~\ref{fig:sun_panel}, using an ephemeris-based model~\cite{michalsky1988astronomical}. The solar zenith angle is
\begin{equation}
  \theta_{z,i} = 90^\circ - \alpha_i.
\end{equation}
The relative air mass $m_i$ representing the irradiation path is defined by
\begin{equation}
  m_i = f_{\mathrm{AM}}(\theta_{z,i}),
\end{equation}
with the Kasten--Young model:
\begin{equation}
\begin{aligned}
  m_i = &
  \left(
    \cos \theta_{z,i}
    +
    0.50572 \,
    (96.07995^\circ - \theta_{z,i})^{-1.6364}
  \right)^{-1}, \\
  &  \theta_{z,i} < 90^\circ .
\end{aligned}
\end{equation}

For the calculation of solar power profiles, we retrieve typical meteorological year (TMY) data for the location from open sources, using:
\begin{itemize}
    \item $G_{b,n,i}$: direct beam irradiance normal to the Sun (W/m$^2$),
    \item $G_{h,i}$: global horizontal irradiance (W/m$^2$),
    \item $G_{d,h,i}$: diffuse horizontal irradiance (W/m$^2$),
    \item $T_{\mathrm{amb},i}$: ambient air temperature at 2\,m (°C),
    \item $v_{\mathrm{w},i}$: wind speed at 10\,m (m/s),
\end{itemize}
aligned with $\{t_i\}_{i=1}^{N}$. For a PV array with tilt $\beta$ and azimuth $\gamma_p$, we use the Perez transposition model~\cite{perez1990modeling} to obtain plane‑of‑array (POA) irradiance components:
\begin{equation}
\begin{aligned}
  &\bigl(
    G_{\mathrm{POA,dn},i},\,
    G_{\mathrm{POA,diff},i},\,
    G_{\mathrm{POA,refl},i}
  \bigr) \\
  &= f_{\mathrm{POA}}\!\bigl(
    \beta, \gamma_p,\,
    \theta_{z,i}, \gamma_{s,i},\,
    G_{b,n,i}, G_{h,i}, G_{d,h,i}
  \bigr),
\end{aligned}
\end{equation}
with total POA irradiance
\begin{equation}
  G_{\mathrm{POA},i}
  =
  G_{\mathrm{POA,dn},i}
  +
  G_{\mathrm{POA,diff},i}
  +
  G_{\mathrm{POA,refl},i}.
\end{equation}

We estimate the PV module temperature using the Faiman model~\cite{faiman2008assessing}:
\begin{equation}
  T_{\mathrm{PV},i}
  =
  T_{\mathrm{amb},i}
  +
  \frac{
    G_{\mathrm{POA},i}
  }{
    U_0 + U_1 \, v_{\mathrm{w},i}
  },
\end{equation}
where $U_0$ (W/m$^2$K) and $U_1$ (W/m$^3$K) are empirical heat loss coefficients. We employ the analytical derate (ADR) model~\cite{king2004photovoltaic} to capture efficiency dependence on irradiance and temperature. It yields PV array efficiency $\eta_{\mathrm{PV},i}$:
\begin{equation}
  \eta_{\mathrm{PV},i}
  =
  f_{\mathrm{ADR}}\!\left(
    G_{\mathrm{POA},i},
    T_{\mathrm{PV},i};
    k_a, k_d, t_{c,d}, k_{rs}, k_{rsh}
  \right),
\end{equation}
where in this work, we use the following parameters
\begin{equation}
\begin{aligned}
  k_a     &= 0.99924, \\
  k_d     &= -5.49097, \\
  t_{c,d} &= 0.01918, \\
  k_{rs}  &= 0.06999, \\
  k_{rsh} &= 0.26144.
\end{aligned}
\end{equation}
The analytical form of $f_{\mathrm{ADR}}$ is available in~\cite{king2004photovoltaic}.

To obtain DC power output, we scale by the nominal array power at standard test conditions (STC), $P_{\mathrm{STC}}$, and the ratio of POA irradiance to STC irradiance $G_{\mathrm{STC}}$:
\begin{equation}
  P_{\mathrm{PV},i}
  =
  P_{\mathrm{STC}} \,
  \eta_{\mathrm{PV},i} \,
  \frac{G_{\mathrm{POA},i}}{G_{\mathrm{STC}}}.
\end{equation}
We take $P_{\mathrm{STC}}$ to approximately correspond to 1\,m$^2$ of panel. The annual hourly PV power profile is
\begin{equation}
  \bigl\{
    (t_i, P_{\mathrm{PV},i})
  \bigr\}_{i=1}^{365 \times 24},
\end{equation}
usable as a synthetic generation profile for detected panels in area of interest.

\section{Numerical assessment of the proposed solar power profiling}\label{sec:evaluation}

\begin{table}[!t]
    \centering
    \begin{threeparttable}
    \caption{Configurations of foundation AI vision models.}
    \label{tab:AI_model_configurations}
    
    \begin{tabular}{ccc}
        \hline
        Model name & \tabincell{c}{Amount of model\\ parameters} & \tabincell{c}{Release date} \\
        \hline
        SAM3\tnote{1} & 859,921,848 & Nov. 19, 2025\\
        \hline
        SAM3.1\tnote{2} & 848,179,510 & Mar. 26, 2026\\
        \hline
    \end{tabular}

    \begin{tablenotes}
      \item[1] \url{https://huggingface.co/facebook/sam3}, accessed 2026-04-26.
      \item[2] \url{https://huggingface.co/facebook/sam3.1}, accessed 2026-04-26.
    \end{tablenotes}
    \end{threeparttable}
\end{table}

We test the approach in Berg am Irchel and Bülach both in Switzerland, which is covered by open and high-resolution (5-cm) aerial images from JOSM. We extract building GIS geometries from OSM and retrieve aerial images from JOSM\footnote{We use the aerial imagery for Canton Zurich in Switzerland at \url{https://josm.openstreetmap.de/mapsview?entry=Canton\%20Zurich\%2C\%20Orthophoto\%20ZH\%20Spring\%202021\%20RGB\%205cm}, accessed 2026-04-26.}. We keep both the building aerial image and building mask at 1500$\times$1500 pixels with zoom level 21. For the detection, we use two openly released foundation AI models, SAM3 and SAM3.1, shown in Table~\ref{tab:AI_model_configurations} with a 70\% confidence threshold selected from initial tests. We assume a fixed panel tilt of 180 degrees (south-facing) and a panel azimuth equal to the city latitude. We set $P_{\mathrm{STC}}=200\,W$ (roughly 1\,m$^2$ of panel) and $G_{\mathrm{STC}}=1000\, W/m^2$. We retrieve TMY data from PVGIS\footnote{\url{https://re.jrc.ec.europa.eu/pvg_tools/en/}, accessed 2026-04-26.}. We release the software, API, and detection results at our Github page\footnote{\url{https://github.com/slzhang-git/GeoSolarDataAI}, accessed 2026-04-26.}. All evaluations are run in Google Colab with T4 GPU. 

\begin{table*}
    \centering
    \caption{Solar panels in the area of interest.}
    \label{tab:areas}
    \begin{tabular}{ccccccc}
        \hline
        Place & \tabincell{c}{Total area (m$^2$)} & \tabincell{c}{Number of\\ buildings} & \tabincell{c}{Area covered\\ by buildings (m$^2$)} & \tabincell{c}{Number (and \\area in m$^2$) of \\labeled solar\\ panels by OSM} & \tabincell{c}{Number (and\\ area in m$^2$) of\\ solar panels\\ detected by\\ sam3} & \tabincell{c}{Number (and\\ area in m$^2$) of\\ solar panels\\ detected by\\ sam3.1}  \\
        \hline
        Berg am Irchel, Switzerland & 7,007,187.59 & 284 & 64,998.72 & 0 (0) &  84 (2,871.34) &  93 (3,396.18)  \\
        \hline
        Bülach, Switzerland & 16,078,514.69 & 2,828 & 872,602.83 & 6 (28.65) & 1342 (46,468.82)  &  1433 (48,665.36)  \\
        \hline
    \end{tabular}
\end{table*}

\begin{figure}[htbp]
    \centering
    \includegraphics[width=0.9\linewidth]{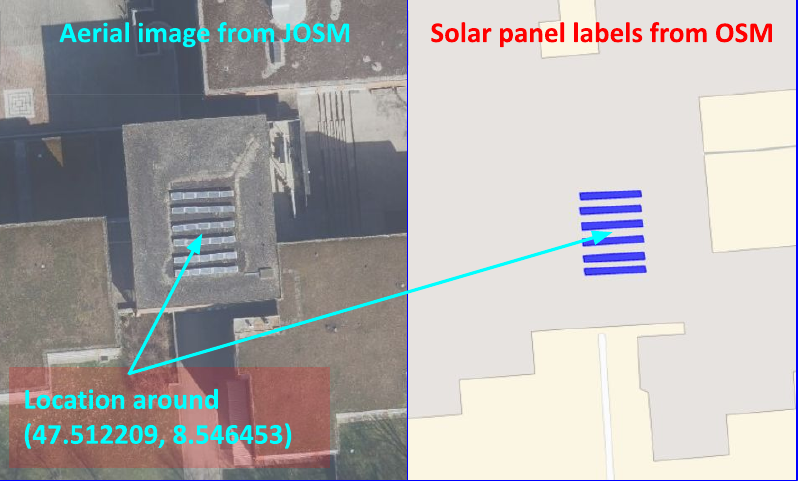}
    \caption{OSM inventory of solar panels for Bülach. Only six panels (right) are publicly available in OSM. For Berg am Irchel, no panels are labeled, indicating very limited OSM coverage for this area.}
    \label{fig:visual_OSM_solar}
\end{figure}

\begin{figure}[htbp]
    \centering
    \includegraphics[width=0.9\linewidth]{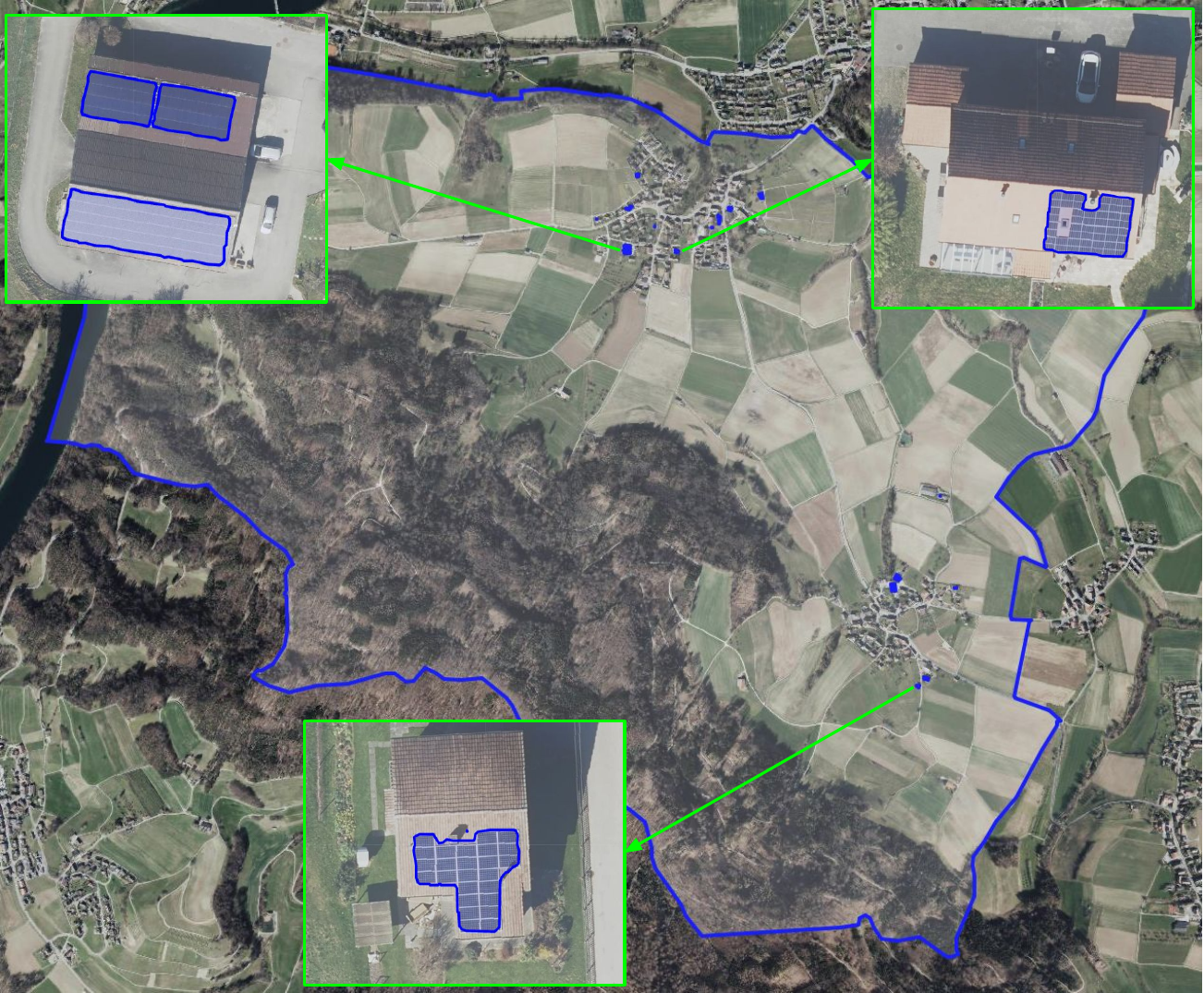}
    \caption{Cleaned solar panel detection result by SAM3 for Berg am Irchel, Switzerland. Interactive map: \url{https://slzhang-git.github.io/GeoSolarDataAI/City_town_level_detection/Cleaned_Berg_Switzerland20260424.html}, accessed 2026-04-26.}
    \label{fig:visual_area1}
\end{figure}

Table~\ref{tab:areas} summarizes detection results. As shown in Figure~\ref{fig:visual_OSM_solar}, OSM contains only six labeled solar panels in Bülach and none in Berg am Irchel. From Table~\ref{tab:areas}, it is clear that our detection substantially extends the solar panel inventory beyond OSM. The newer model (SAM3.1), though with fewer parameters, detects more panels than SAM3, indicating the progress in foundation AI vision models.

\begin{figure}[htbp]
    \centering
    \includegraphics[width=0.9\linewidth]{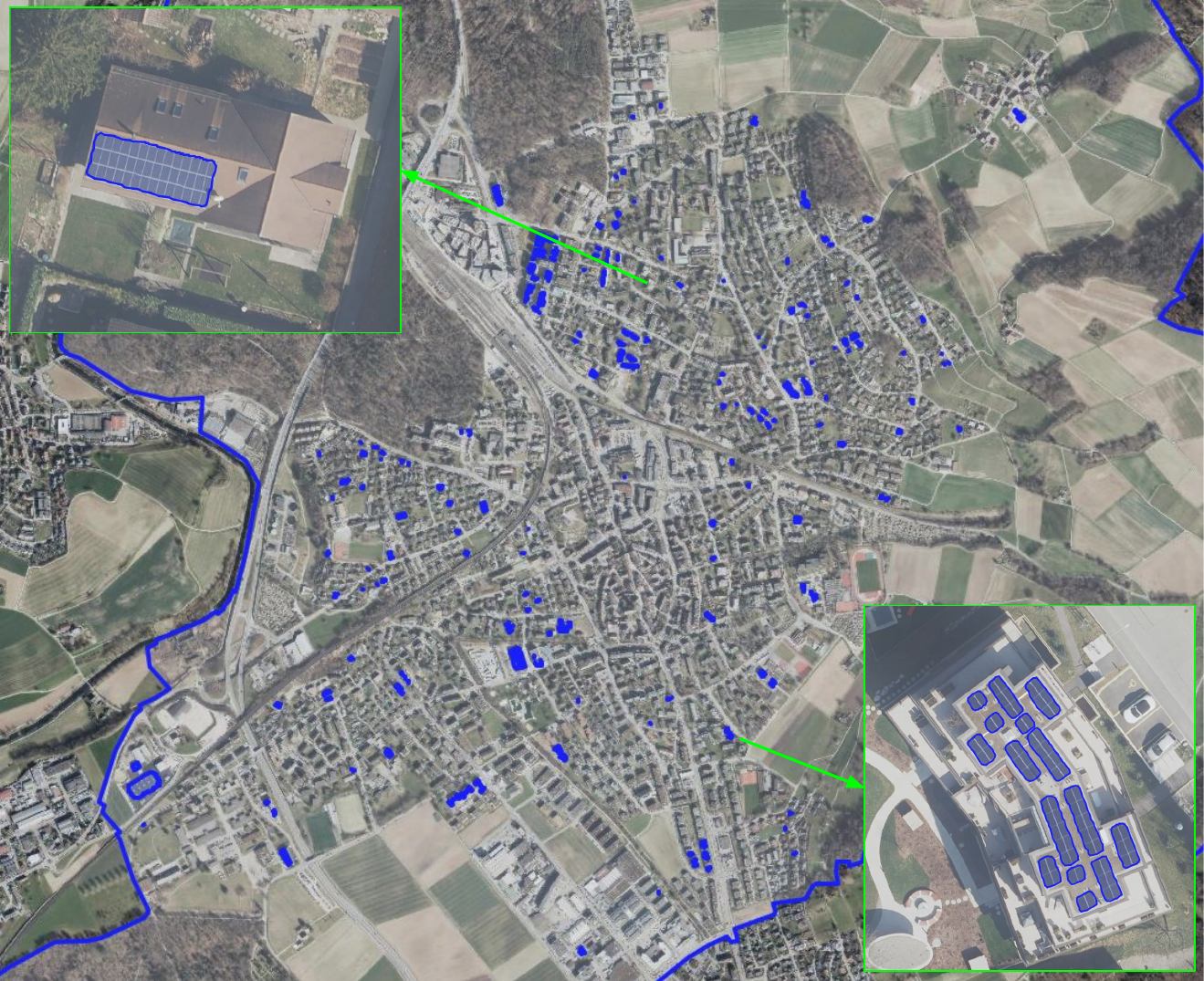}
    \caption{Cleaned solar panel detection result by SAM3 for Bülach, Switzerland. Interactive map: \url{https://slzhang-git.github.io/GeoSolarDataAI/City_town_level_detection/Cleaned_Bulach_Switzerland20260426.html}, accessed 2026-04-26.}
    \label{fig:visual_area2}
\end{figure}

We would like to note that detection errors by AI models are inevitable. To improve data quality, we clean and remove incorrectly detected panels using Geojson.io\footnote{\url{https://geojson.io}, accessed 2026-04-26.}. This introduces limited manual work but yields city-scale solar panel maps from open data. A higher detection threshold can reduce manual cleaning, yet at the risk of more missed detections. After cleaning, we obtain 14 panels (956.29\,m$^2$) in Berg am Irchel and 507 panels (18,947.17\,m$^2$) in Bülach. We show the cleaned solar panel geometries in Figures~\ref{fig:visual_area1} and~\ref{fig:visual_area2}. The raw georeferenced data and interactive maps are accessible via our Github page.

\begin{figure}[htbp]
    \centering
    \includegraphics[width=0.9\linewidth]{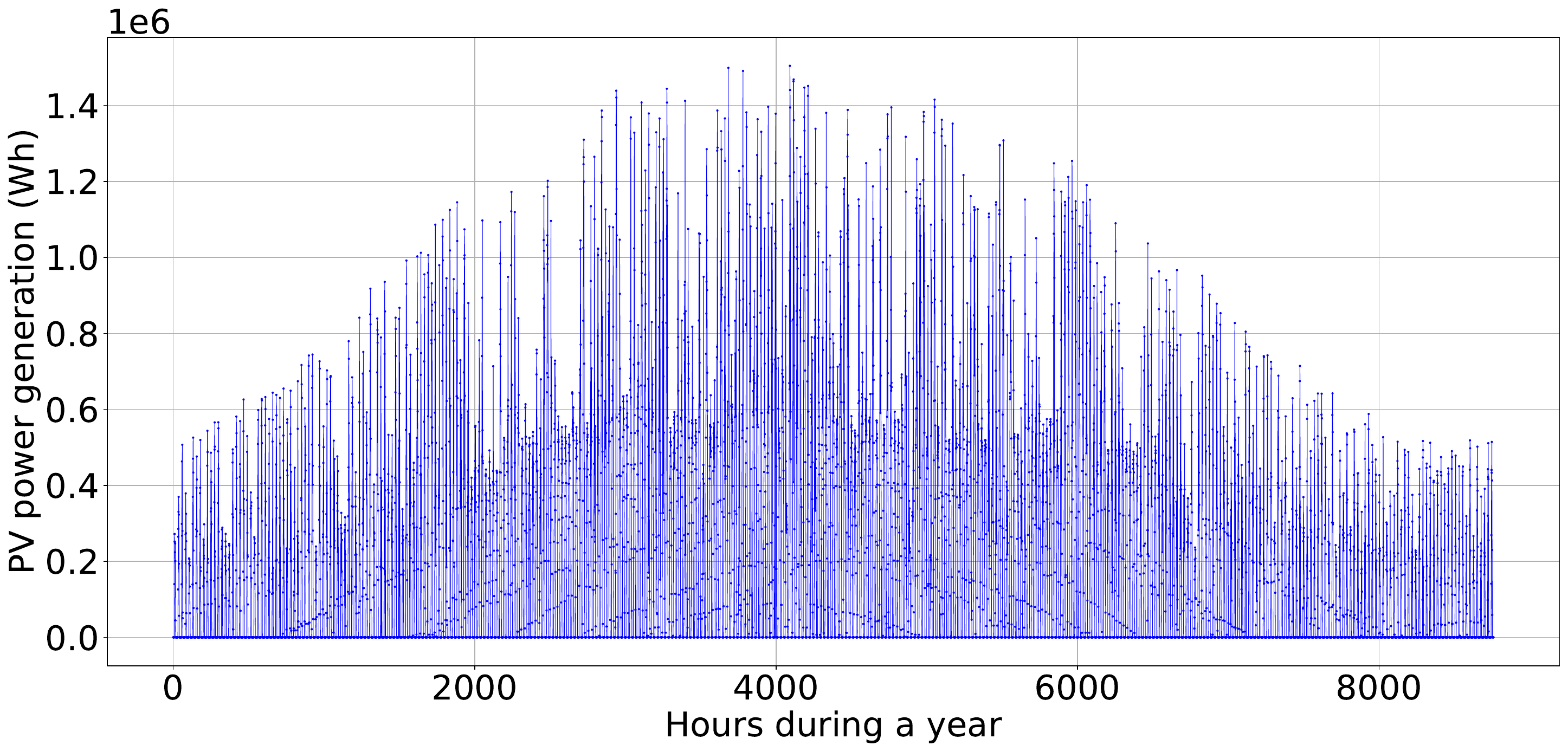}
    \caption{Example yearly solar power profile (hourly resolution) for the city of Bülach. The raw profile data (also for Berg am Irchel) in CSV format is available via our Github page at \url{https://github.com/slzhang-git/GeoSolarDataAI/tree/main/City_town_level_detection}, accessed 2026-04-26.}
    \label{fig:visual_profile_varying_detection_threshold}
\end{figure}

Using the cleaned detection results, we generate solar power profiles for a typical meteorological year for Bülach, shown in Figure~\ref{fig:visual_profile_varying_detection_threshold}. Despite simplified assumptions on weather and panel configurations, the provided profile can be used for researchers in distribution grid power flow analysis and edge capacity estimation. It can play a crucial role for energy planners in estimating needed measures and capacity in integrating the local solar production into the utility grid. Furthermore, it can be used by renewable companies and stakeholders to evaluate the market potential and develop new business models for the participatory energy system.

\section{Conclusions}\label{conclusions}

This paper presented a scalable framework for rooftop solar panel detection and city-level solar power profiling. Our design leverages openly accessible aerial imagery, generalizable AI vision models, and open geographic, and meteorological data. The approach avoids case‑specific model training and proprietary datasets, thus promoting transparency and reproducibility. We converted solar panel pixels detected by AI into georeferenced polygons aligned with OSM. This yields spatially explicit, reusable representations that can be accumulated into an open solar panel inventory, resulting in a scalable framework that works across areas. Using open weather data, we derived time‑varying city-level solar power profiles, linking visual detection with quantitative energy analysis. We released an API that automates building‑level solar panel detection. The API provides georeferenced rooftop solar panel polygons for a building location, which empowers scanning larger areas and generating power profiles. We envision future work on benchmarking different AI models and open imagery providers, so as to identify performance limits and improve the accuracy especially for the solar panel detection in our approach. We also recommend the study to extract granular solar panel status from open data, \textit{e.g.}, the tilt and azimuth for each individual panel, to promote accuracy of the solar power profiling.

\bibliographystyle{unsrt}
\bibliography{bibfile}

@ARTICLE{11364211,
  author={Zhang, Shiliang and Maharjan, Sabita and Strunz, Kai and Bryne, Jan Christian},
  journal={IEEE Data Descriptions}, 
  title={Descriptor: Norwegian Electricity in Geographic Dataset (NoreGeo)}, 
  year={2026},
  volume={3},
  number={},
  pages={82-92},
  doi={10.1109/IEEEDATA.2026.3658039}
  }

@article{li2025global,
  title={Global photovoltaic solar panel dataset from 2019 to 2022},
  author={Li, Anqi and Liu, Luling and Li, Shijie and Cui, Xihong and Chen, Xuehong and Cao, Xin},
  journal={Scientific Data},
  volume={12},
  number={1},
  pages={637},
  year={2025},
  publisher={Nature Publishing Group UK London}
}

@article{li2025transformer,
  title={Transformer approach to nowcasting solar energy using geostationary satellite data},
  author={Li, Ruohan and Wang, Dongdong and Wang, Zhihao and Liang, Shunlin and Li, Zhanqing and Xie, Yiqun and He, Jiena},
  journal={Applied Energy},
  volume={377},
  pages={124387},
  year={2025},
  publisher={Elsevier}
}

@article{awadallah2025road,
  title={The Road to FAIR Energy Data: Motivations, Shortcomings, and Use Cases},
  author={Awadallah, Rawia and Tammaro, Anna Maria},
  journal={International Information \& Library Review},
  volume={57},
  number={2},
  pages={177--192},
  year={2025},
  publisher={Taylor \& Francis}
}

@article{nyangiwe2026performance,
  title={Performance Monitoring of Photovoltaic Modules Using Machine-Learning-Based Solutions: A Survey of Current Trends},
  author={Nyangiwe, Nangamso Nathaniel and Kenfack, Abraham Dimitri Kapim and Thantsha, Nicolas and Msimanga, Mandla},
  journal={Energy Science \& Engineering},
  volume={14},
  number={3},
  pages={1663--1682},
  year={2026},
  publisher={Wiley Online Library}
}

@article{barraz2025fast,
  title={Fast and automatic solar module geo-labeling for optimized large-scale photovoltaic systems inspection from UAV thermal imagery using deep learning segmentation},
  author={Barraz, Zoubir and Sebari, Imane and Lamrini, Nassim and Ait El Kadi, Kenza and Ait Abdelmoula, Ibtihal},
  journal={Cleaner Engineering and Technology},
  pages={101048},
  year={2025},
  publisher={Elsevier}
}

@article{guo2025solar,
  title={Solar photovoltaic assessment with large language model},
  author={Guo, Muhao and Weng, Yang},
  journal={Applied Energy},
  volume={402},
  pages={126835},
  year={2025},
  publisher={Elsevier}
}

@article{boccalatte2025leveraging,
  title={Leveraging large-scale aerial data for accurate urban rooftop solar potential estimation via multitask learning},
  author={Boccalatte, Alessia and Jha, Ankit and Chanussot, Jocelyn},
  journal={Solar Energy},
  volume={290},
  pages={113336},
  year={2025},
  publisher={Elsevier}
}

@article{amorim2025collaborative,
  title={Collaborative inspection of solar panel farms using YOLOv5-based computer vision and UGV-UAV integration},
  author={Amorim, Johann SJC and Neto, Accacio FS and Chaves, Rafael S and Zachi, Alessandro RL and Gouv{\^e}a, Josiel A and Andrade, Fabio AA and Pinto, Milena F},
  journal={Journal of Intelligent \& Robotic Systems},
  volume={111},
  number={2},
  pages={66},
  year={2025},
  publisher={Springer}
}

@article{mandourah2025analysing,
  title={Analysing the use of OpenAerialMap images for OpenStreetMap edits},
  author={Mandourah, Ammar and Hochmair, Hartwig H},
  journal={Geo-spatial Information Science},
  volume={28},
  number={3},
  pages={1179--1194},
  year={2025},
  publisher={Taylor \& Francis}
}

@inproceedings{deri2025crowdsourced,
  title={Crowdsourced Data for Urban Planning: A Critical Evaluation of OpenStreetMap Accuracy and Completeness},
  author={Deri, Federica and Mara, Federico and Anselmi, Chiara},
  booktitle={International Conference on Computational Science and Its Applications},
  pages={403--420},
  year={2025},
  organization={Springer}
}

@article{rao2025development,
  title={Development of a smart cloud-based monitoring system for solar photovoltaic energy generation},
  author={Rao, Challa Krishna and Sahoo, Sarat Kumar and Yanine, Franco Fernando},
  journal={Unconventional Resources},
  volume={6},
  pages={100173},
  year={2025},
  publisher={Elsevier}
}

@article{cardoso2024automated,
  title={Automated detection and tracking of photovoltaic modules from 3D remote sensing data},
  author={Cardoso, Andressa and Jurado-Rodr{\'\i}guez, David and L{\'o}pez, Alfonso and Ramos, M Isabel and Jurado, Juan Manuel},
  journal={Applied Energy},
  volume={367},
  pages={123242},
  year={2024},
  publisher={Elsevier}
}

@article{sezer2021detection,
  title={Detection of solder paste defects with an optimization-based deep learning model using image processing techniques},
  author={Sezer, Ali and Altan, Ayta{\c{c}}},
  journal={Soldering \& Surface Mount Technology},
  volume={33},
  number={5},
  pages={291--298},
  year={2021},
  publisher={Emerald Publishing Limited}
}

@article{park2023boost,
  title={Boost-up efficiency of defective solar panel detection with pre-trained attention recycling},
  author={Park, YeongHyeon and Kim, Myung Jin and Gim, Uju and Yi, Juneho},
  journal={IEEE Transactions on Industry Applications},
  volume={59},
  number={3},
  pages={3110--3120},
  year={2023},
  publisher={IEEE}
}

@article{yang2025large,
  title={A large-scale ultra-high-resolution segmentation dataset augmentation framework for photovoltaic panels in photovoltaic power plants based on priori knowledge},
  author={Yang, Ruiqing and He, Guojin and Yin, Ranyu and Wang, Guizhou and Peng, Xueli and Zhang, Zhaoming and Long, Tengfei and Peng, Yan and Wang, Jianping},
  journal={Applied Energy},
  volume={390},
  pages={125879},
  year={2025},
  publisher={Elsevier}
}

@article{10.1145/3711118,
author = {Zha, Daochen and Bhat, Zaid Pervaiz and Lai, Kwei-Herng and Yang, Fan and Jiang, Zhimeng and Zhong, Shaochen and Hu, Xia},
title = {Data-centric Artificial Intelligence: A Survey},
year = {2025},
issue_date = {May 2025},
publisher = {Association for Computing Machinery},
address = {New York, NY, USA},
volume = {57},
number = {5},
url = {https://doi.org/10.1145/3711118},
journal = {ACM Comput. Surv.},
month = jan,
articleno = {129},
numpages = {42}
}

@article{bradbury2016distributed,
  title={Distributed solar photovoltaic array location and extent dataset for remote sensing object identification},
  author={Bradbury, Kyle and Saboo, Raghav and L Johnson, Timothy and Malof, Jordan M and Devarajan, Arjun and Zhang, Wuming and M Collins, Leslie and G Newell, Richard},
  journal={Scientific data},
  volume={3},
  number={1},
  pages={160106},
  year={2016},
  publisher={Nature Publishing Group}
}

@InProceedings{pmlr-v267-shabbir25a,
  title = 	 {{G}eo{P}ixel: Pixel Grounding Large Multimodal Model in Remote Sensing},
  author =       {Shabbir, Akashah and Zumri, Mohammed and Bennamoun, Mohammed and Khan, Fahad Shahbaz and Khan, Salman},
  booktitle = 	 {Proceedings of the 42nd International Conference on Machine Learning},
  pages = 	 {54095--54111},
  year = 	 {2025},
  volume = 	 {267},
  series = 	 {Proceedings of Machine Learning Research},
  month = 	 {13--19 Jul},
  publisher =    {PMLR},
  url = 	 {https://proceedings.mlr.press/v267/shabbir25a.html}
}

@misc{hu2026extracting,
  title={Extracting and analysing geographic information from natural language texts},
  author={Hu, Xuke and Purves, Ross S and Moncla, Ludovic and Kersten, Jens and Stock, Kristin},
  journal={International Journal of Geographical Information Science},
  volume={40},
  number={3},
  pages={631--644},
  year={2026},
  publisher={Taylor \& Francis}
}

@article{perez1990modeling,
  title={Modeling daylight availability and irradiance components from direct and global irradiance},
  author={Perez, Richard and Ineichen, Pierre and Seals, Robert and Michalsky, Joseph and Stewart, Ronald},
  journal={Solar energy},
  volume={44},
  number={5},
  pages={271--289},
  year={1990},
  publisher={Elsevier}
}

@article{faiman2008assessing,
  title={Assessing the outdoor operating temperature of photovoltaic modules},
  author={Faiman, David},
  journal={Progress in Photovoltaics: Research and Applications},
  volume={16},
  number={4},
  pages={307--315},
  year={2008},
  publisher={Wiley Online Library}
}

@book{king2004photovoltaic,
  title={Photovoltaic array performance model},
  author={King, David L and Kratochvil, Jay A and Boyson, William Earl},
  volume={8},
  year={2004},
  publisher={United States. Department of Energy}
}

@article{michalsky1988astronomical,
  title={The astronomical almanac's algorithm for approximate solar position (1950--2050)},
  author={Michalsky, Joseph J},
  journal={Solar energy},
  volume={40},
  number={3},
  pages={227--235},
  year={1988},
  publisher={Elsevier}
}

\end{document}